\documentclass[conference]{IEEEtran}
\IEEEoverridecommandlockouts
\usepackage{cite}
\usepackage{amsmath,amssymb,amsfonts}
\usepackage{algorithmic}
\usepackage{graphicx}
\usepackage{textcomp}
\def\BibTeX{{\rm B\kern-.05em{\sc i\kern-.025em b}\kern-.08em
    T\kern-.1667em\lower.7ex\hbox{E}\kern-.125emX}}

\usepackage{booktabs}       % professional-quality tables
\usepackage{nicefrac}       % compact symbols for 1/2, etc.
\usepackage{microtype}      % microtypography
\usepackage{lipsum}		% Can be removed after putting your text content
\usepackage{doi}
\usepackage{subfiles}
\usepackage[linesnumbered,lined,boxed,commentsnumbered]{algorithm2e}
\usepackage{caption}
\usepackage{subcaption}
\usepackage{cancel}
\usepackage{cuted}
\usepackage{booktabs}
\usepackage[table,xcdraw]{xcolor}
\usepackage{tabularray}

\begin{document}

\title{DouRN: Improving DouZero by Residual Neural Networks}

\author{\IEEEauthorblockN{Yiquan Chen}
	\IEEEauthorblockA{\textit{School of AI and Advanced Computing} \\
		\textit{Xi'an Jiaotong-Liverpool University}\\
		Suzhou, China \\
		Yiquan.Chen19@student.xjtlu.edu.cn}
	\and
	\IEEEauthorblockN{Yingchao Lyu}
	\IEEEauthorblockA{\textit{School of AI and Advanced Computing} \\
	\textit{Xi'an Jiaotong-Liverpool University}\\
	Suzhou, China \\
	Yingchao.Lyu@student.xjtlu.edu.cn}
	\and
	\IEEEauthorblockN{Di Zhang*}
	\IEEEauthorblockA{\textit{School of AI and Advanced Computing} \\
	\textit{Xi'an Jiaotong-Liverpool University}\\
	Suzhou, China \\
	Di.Zhang@student.xjtlu.edu.cn}
}

\maketitle

\begin{abstract}
Deep reinforcement learning has made significant progress in games with imperfect information, but its performance in the card game Doudizhu (Chinese Poker/Fight the Landlord) remains unsatisfactory. Doudizhu is different from conventional games as it involves three players and combines elements of cooperation and confrontation, resulting in a large state and action space. In 2021, a Doudizhu program called DouZero\cite{zha2021douzero} surpassed previous models without prior knowledge by utilizing traditional Monte Carlo methods and multilayer perceptrons. Building on this work, our study incorporates residual networks into the model, explores different architectural designs, and conducts multi-role testing. Our findings demonstrate that this model significantly improves the winning rate within the same training time. Additionally, we introduce a call scoring system to assist the agent in deciding whether to become a landlord. With these enhancements, our model consistently outperforms the existing version of DouZero and even experienced human players. \footnote{The source code is available at \url{https://github.com/Yingchaol/Douzero_Resnet.git.}} 
\end{abstract}


\section{Introduction}

The objective of this study is to develop a deep reinforcement learning-based model for DouDiZhu, a card game that presents unique challenges when applying reinforcement learning techniques from other games. One key challenge is that DouDiZhu involves three players, namely a landlord and two peasants, where the peasants collaborate against the landlord, resulting in a game that combines both cooperation and competition. Consequently, the widely used counterfactual regret minimization algorithm employed in poker\cite{brown2019superhuman} cannot be directly applied to this complex three-player games. 

Furthermore, the set of permissible card combinations changes dynamically as the game progresses. In contrast to Texas Hold'em\cite{brown2019superhuman}, where Deep Reinforcement Learning (DRL) has been successfully applied, DouDiZhu encompasses a large action space that is difficult to explore. Consequently, traditional reinforcement learning algorithms suffer from high search costs and prove ineffective. Both the Deep Q-Learning and Asynchronous Advantage Actor-Critic (A3C) algorithms\cite{mnih2016asynchronous}, well-established in the field of DRL, are unsuitable for DouDiZhu training due to these reasons. Deep Q-Learning is prone to overestimation issues in large action spaces, significantly impacting training outcomes, while A3C struggles to fully exploit the features of the action space, limiting its ability to generalize to unknown actions\cite{dulac2015deep}. According to \cite{you2020combinatorial}, both models exhibit remarkably low winning rates against simple rule-based models. 

Although deep learning is still necessary, training DouDiZhu using MLPs presents challenges such as tuning difficulties, extensive training time, and slow learning speed, as observed in DouZero\cite{zha2021douzero}. To address these issues, we adopted the residual neural network (ResNet) to enhance the original multilayer perceptrons (MLPs) in DouZero, which offers potential solutions to the degradation problem of deep neural networks and facilitates faster convergence\cite{zhang2019deep}. 

Here we introduced a simple multi-layer perceptron based on a call-score network to enable the AI to determine whether it can play as a landlord and obtain three additional hidden cards. When opportunity comes, becoming a landlord not only increases the probability of winning but also provides a double score bonus. Accurately assessing the score at the beginning of each game by analyzing one's own cards and the scores given by opponents is crucial. 

By combining these two enhancements, the newly trained model, DouRN, outperforms the original DouZero model and exhibits superior capability against experienced DouDiZhu players. 

\section{Related Work}

There exists a substantial body of literature pertaining to the application of DRL for DouDiZhu. Notably, a recent study\cite{you2020combinatorial} introduced a Combinational Q-Learning algorithm that combines a reduced action space and maximum pooling technique, resulting in promising outcomes for DouDiZhu gameplay. Furthermore, paper \cite{jiang2019deltadou} proposed a model named DeltaDou, which integrates Monte Carlo tree search with DRL. However, this model exhibits limitations in terms of computational cost and reliance on human a priori knowledge, thereby constraining its performance and generalizability. In contrast, the DouZero model, which employs pure DRL, has outperformed all previous AI models in DouDiZhu gameplay\cite{zha2021douzero}. Building upon this, DouZero+\cite{zhao2022douzero+} have enhanced the DouZero model through opponent modeling and coach-guided learning. 

ResNet is a deep learning architecture that revolutionized image classification tasks. It addresses the problem of vanishing gradients in deep networks by introducing skip connections, also known as residual connections. These connections allow the network to directly propagate information from earlier layers to later ones, enabling faster and more accurate learning. This architecture has significantly improved the training of deep neural networks, leading to state-of-the-art performance in various computer vision tasks\cite{zhang2019deep}. However, the application of ResNet in DRL has only recently gained attention. For instance, a DRL model named FullyConv, designed for StarCraft II, incorporates ResNet into its architecture\cite{vinyals2019grandmaster}. Furthermore, paper\cite{wang2019novel} successfully utilized ResNet for deep learning in Chinese Mahjong, a complex incomplete information game similar to DouDiZhu. Comparative analysis against conventional approaches revealed that ResNet can achieve better winning rate and score, indicating its superior performance. 

\section{Methodology}
\subsection{Residual Networks}\label{sec:rn}

\begin{figure}
	\centering
	\includegraphics[width=1\linewidth]{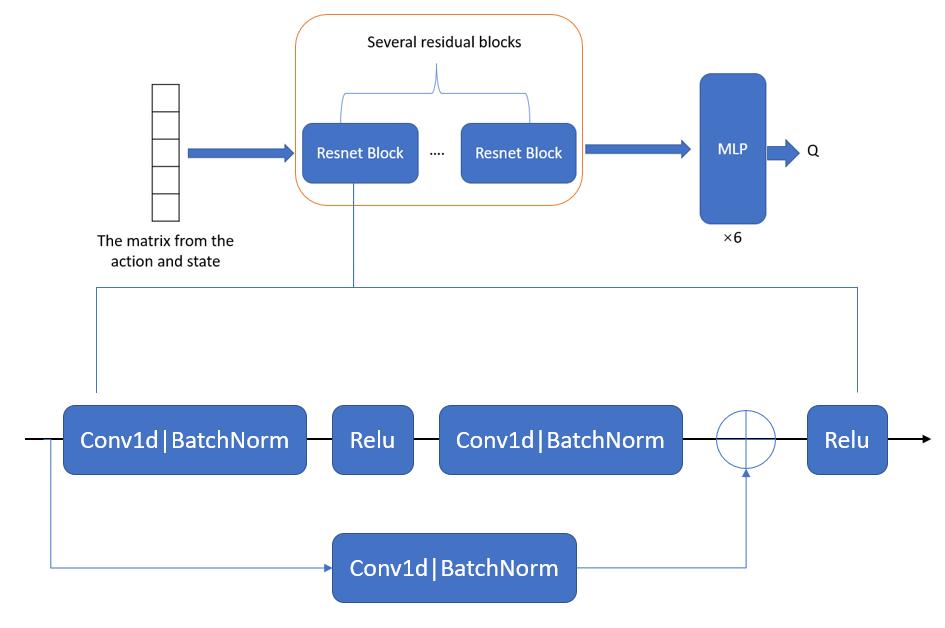}
	\caption{The architecture of a residual block.}
	\label{fig:11}
\end{figure}

The key to designing a residual neural network is to build a residual block, which usually consists of multiple convolutional layers and batch normalization layers\cite{zhang2019deep}. In the residual block, the input of each convolutional layer is the output of the previous layer plus a skip connection (i.e., residual connection). The fundamental structure of a residual block we used here is illustrated in Fig. \ref{fig:11}. The original neural network structure of DouZero is a 6-layer MLP. To optimize the network, several residual blocks can be appended. Generally, the use of residual blocks may enhance algorithmic training effectiveness, but deepening neural networks may also increases the consumption of computing resources. Therefore, we need to carefully design the network architecture.

\begin{figure}
	\centering
	\includegraphics[width=1\linewidth]{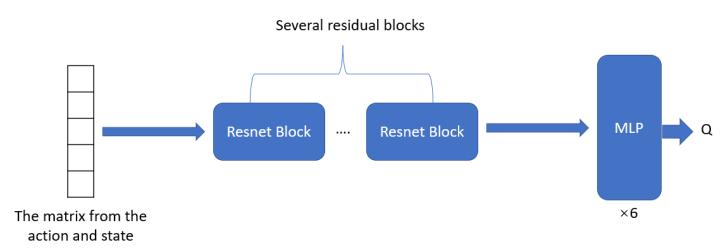}
	\caption{Insert several residual blocks before the MLP.}
	\label{fig:1a}
\end{figure}

\begin{figure}
	\centering
	\includegraphics[width=1\linewidth]{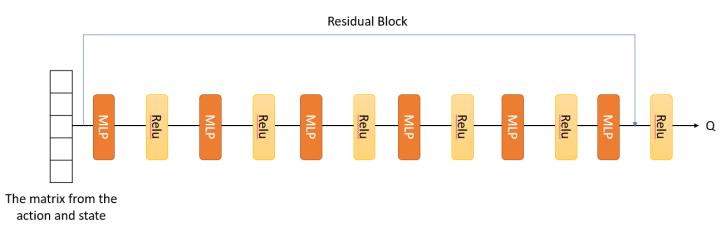}
	\caption{Replace the original 6-layer MLP into residual blocks.}
	\label{fig:1b}
\end{figure}

We have designed two architectures. In the first architecture (Fig. \ref{fig:1a}), we directly stack residual blocks on top of the MLP in the original DouZero, while keeping the training method unchanged. This allows us to observe the impact of the residual blocks. After achieving positive effects with fewer stacked layers, we can continuously improve the model performance by increasing the number of residual blocks. In the second architecture (Fig. \ref{fig:1b}), we replace the entire MLP in the original DouZero with residual blocks while maintaining the overall number of layers. We hope that this approach can minimize the increase in model training time. 

%The first architecture involves adding a residual block for training and training a pure MLP model within the same time frame (Fig. \ref{fig:1a}). 

%The models for the landlord and peasants are compared to analyze the impact of adding the residual block. Based on the results, more residual blocks are added, and the aforementioned steps are repeated until the optimal neural network architecture is achieved. 

%Alternatively, the original MLP network can be reconstructed into a residual network without adding extra residual blocks, where residuals are incorporated into the MLP (Fig. \ref{fig:1b}). This optim00ization method is expected to have a minimal impact on training time. 

\subsection{Bidding System}\label{sec:bs}

After the game starts, each player has 17 cards in his hand, and the remaining 3 cards are hidden cards. The first player to be called Landlord is determined randomly, usually automatically selected by the system. Other players take turns calling the landlord, and they can choose to call the landlord or not. When bidding on a landowner, he must offer a higher multiplier (1, 2 or 3) than the previous player. The hidden card will be taken by the highest bidder and added to the hand. The landowner will play his cards first, and the other two farmers will become his opponents.

%As the game of DouDiZhu primarily focuses on the player's hand and the scores provided by other players, a straightforward Multi-Layer Perceptron (MLP) model is deemed sufficient for scoring purposes\cite{zhao2022douzero+}. 

\begin{figure}
	\centering
	\includegraphics[width=1\linewidth]{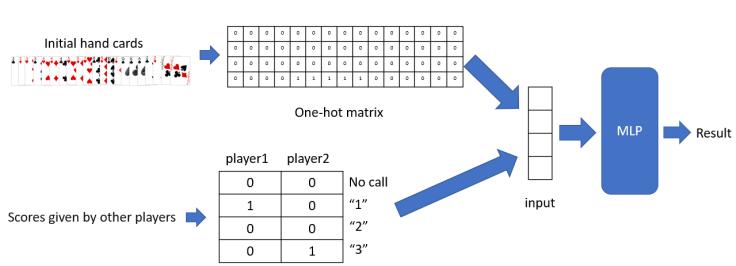}
	\caption{The input data utilized in the bidding system encompasses the player's hand as well as the scores assigned by the two opposing players. Notably, the scores given by the other players are transformed into a one-hot matrix representation for further analysis and processing.}
	\label{fig:2}
\end{figure}

Obviously, a decision-making system needs to be added here, which includes analysis of the current hand and the opponents' bidding multiplier. The potential benefits must outweigh the risks. The player's hand is transformed into a 4×15 matrix representation. Furthermore, the score provided by the player is converted into a 4×2 one-hot matrix representation. For instance, as illustrated in Fig. \ref{fig:2}, player 1 assigns a score of 1, while player 2 assigns a score of 3, resulting in the generation of a matrix representation. 

\begin{figure}
	\centering
	\includegraphics[width=1\linewidth]{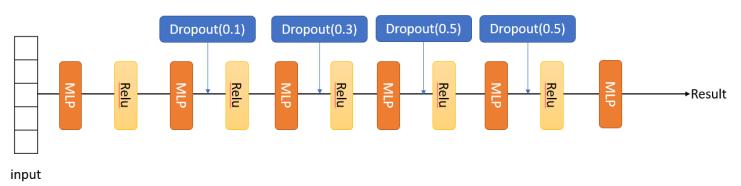}
	\caption{In the architecture of the bidding system, some neurons will be randomly discarded in dropout without training, and the probability is indicated within parentheses.}
	\label{fig:3}
\end{figure}

The MLP architecture of the bidding system is depicted in Fig. \ref{fig:3}. It comprises 6 layers of neural networks, incorporating dropouts to mitigate overfitting and prevent excessive activation of all neurons. The network ultimately produces an output value, which determines the action assigned by the trained model. 

\section{Experiments}

We first evaluate the performance of the two residual network architectures along in Sec. \ref{sec:rn}. To evaluate the performance of the enhanced model\footnote{The experiments were conducted on the Sugon platform, utilizing a server equipped with an 8-core Hygon C86 7185 CPU @ 2.40 GHz, 64 GB of RAM, and two Tesla T4 graphics cards.}, we adopted a model validation approach similar to that employed by DouZero. This involved playing landlords and peasants from different models against each other. Specifically, one model represented the improved DouRN, while the other represented the original DouZero, with each assuming the role of landlord and peasant respectively. Then, their roles were swapped, enabling them to compete against each other using the same deck of cards. The evaluation criteria were based on the winning rate, calculated as the number of games won by the model divided by the total number of games played. 

Furthermore, to assess the bidding system in Sec. \ref{sec:bs}, selected real players were invited to compete against the models, and their performance was evaluated based on their winning percentage. Except for the newly added residual blocks, DouRN is consistent with DouZero in all hyperparameters. This facilitates our comparison.

\subsection{Evaluation on Residual Networks}\label{sec:ern}

\subsubsection*{2 Extra Layers}

\begin{table}
	\centering
	\caption{Comparison of a DouRN model trained with a 2-layer residual block and a DouZero model with the different roles and training steps. L. stands for landowner and P. stands for peasant.}
	\begin{tblr}{
			width = \linewidth,
			colspec = {Q[273]Q[246]Q[248]Q[160]},
			vline{2,4} = {-}{},
			hline{1,5} = {-}{0.08em},
			hline{2} = {1-3}{0.03em},
			hline{2} = {4}{},
		}
		{Step(×$10^5$)\\/2 Blocks} & {DouRN(L.)\\ vs. DouZero(P.)} & {DouRN(P.) \\ vs. DouZero(L.)} & {Trainning\\Time (h)} \\
		1.0                        & 49.02\%                       & 51.32\%                        & 56                    \\
		2.0                        & 51.92\%                       & 52.10\%                        & 112                   \\
		3.0                        & 52.02\%                       & 53.77\%                        & 168                   
	\end{tblr}
\end{table}\label{tab:1}

 In order to assess the impact on model performance, the model was initially trained with 2 residual blocks for a total of $3.0\times 10^5$ steps. As indicated in Tab. \ref{tab:1}, it can be observed that although the DouRN model does not outperform DouZero in the early stages of training as a landlord, it slightly surpasses DouZero in terms of winning rate at the end of training, both as a landlord and a peasant. Hence, a preliminary conclusion can be drawn that the inclusion of residual blocks can be beneficial for enhancing the model. 

\subsubsection*{4/6 Extra Layers}

\begin{table}
	\centering
	\caption{Winning rate of match between different residual block models and DouZero models.}
	\begin{tblr}{
			width = \linewidth,
			colspec = {Q[271]Q[242]Q[246]Q[158]},
			vline{2,4} = {-}{},
			hline{1,15} = {-}{0.08em},
			hline{2,8-9} = {1-3}{0.03em},
			hline{2,8-9} = {4}{},
		}
		{Step(×$10^5$)\\/4 Blocks} & {DouRN(L.)\\ vs. DouZero(P.)} & {DouRN(P.)\\ vs. DouZero(L.)} & {Trainning\\Time (h)} \\
		1.0                        & 0.5221                        & 0.5322                        & 68                    \\
		2.0                        & 0.5403                        & 0.5512                        & 136                   \\
		3.0                        & 0.5508                        & 0.5551                        & 204                   \\
		4.0                        & 0.5549                        & 0.5578                        & 272                   \\
		5.0                        & 0.5562                        & 0.5592                        & 340                   \\
		6.0                        & 0.5580                        & 0.5640                        & 408                   \\
		{Step(×$10^5$)\\/6 Blocks} & {DouRN(L.)\\ vs. DouZero(P.)} & {DouRN(P.)\\ vs. DouZero(L.)} & {Trainning\\Time (h)} \\
		1.0                        & 0.5397                        & 0.5410                        & 77                    \\
		2.0                        & 0.5611                        & 0.5562                        & 154                   \\
		3.0                        & 0.5641                        & 0.5609                        & 231                   \\
		4.0                        & 0.5683                        & 0.5702                        & 308                   \\
		5.0                        & 0.5701                        & 0.5704                        & 385                   \\
		6.0                        & 0.5704                        & 0.5703                        & 462                   
	\end{tblr}
\end{table}\label{tab:2}

 At this stage, we extended the number of residual blocks to 4 and 6 and increased the training steps to $6.0\times 10^5$. This ensured a sufficient training of the model. The results presented in Tab. \ref{tab:2} demonstrate a consistent increase in winning rate with the addition of residuals. Specifically, the model with 4 residuals exhibited a significantly higher win rate compared to the model with 2 residuals. However, the model with 6 residuals approached convergence at $6.0\times 10^5$ steps, with a win rate similar to that of the model with 4 residuals. Furthermore, the training cost also increased considerably, with the model with 6 residuals taking approximately twice as long as the model with 2 residuals. Thus, it can be predicted that further addition of residuals would enhance the model's performance but at an unacceptable increase in computational effort. 
 
\subsubsection*{Simple Replacement}

\begin{table}
	\centering
	\caption{Comparison of win rates between the MLP model with residual blocks and the DouZero model.}
	\begin{tblr}{
			width = \linewidth,
			colspec = {Q[271]Q[242]Q[246]Q[158]},
			vline{2,4} = {-}{},
			hline{1,8} = {-}{0.08em},
			hline{2} = {1-3}{0.03em},
			hline{2} = {4}{},
		}
		{Step(×$10^5$)\\/Simple} & {DouRN(L.)\\ vs. DouZero(P.)} & {DouRN(P.) \\ vs. DouZero(L.)} & {Trainning\\Time (h)} \\
		1.0                      & 47.98\%                       & 49.08\%                        & 61                    \\
		2.0                      & 49.11\%                       & 50.96\%                        & 122                   \\
		3.0                      & 49.60\%                       & 50.51\%                        & 183                   \\
		4.0                      & 50.40\%                       & 51.03\%                        & 244                   \\
		5.0                      & 50.50\%                       & 51.23\%                        & 305                   \\
		6.0                      & 50.32\%                       & 50.76\%                        & 366                   
	\end{tblr}
\end{table}\label{tab:4}

The architecture depicted in Fig. \ref{fig:1b} is tested under same procedures as before. Table 4 reveals that the new model exhibits similar winning rates to the DouZero model, and even performs worse at lower step counts. In addition, there is no notable distinction between the two models in terms of training time. Therefore, the effect of simple replacement is not satisfactory.

\subsubsection*{Comparative Analysis}

Networks with additional layers are superior to networks without additional layers in this game, which is probably due to the following reasons: 

ResNet adopts more residual blocks and has a greater depth, which means it can have more levels to learn complex features and explore more strategies and states. This enables ResNet to better capture subtle variations and complex relationships in the experience of play, thereby improving network performance. 

ResNet has many hyperparameters that need to be adjusted, such as network depth and the number of residual blocks. If the MLP in DouZero is simply replaced by a residual network without proper optimization, it may not achieve the desired effect. 

\subsubsection*{Limitation}

Firstly, limited computational resources hindered sufficient training, leading to suboptimal neural network architecture and model performance. Secondly, the model exhibited unwise decisions in certain testing scenarios. Lastly, the inclusion of residual networks significantly reduced training speed. 

%\begin{figure}
%	\centering
%	\includegraphics[width=0.8\linewidth]{12}
%	\caption{Winning rates of DouRN models with different steps and DouZero models that have been trained for 6*1e6 steps.}
%	\label{fig:12}
%\end{figure}
%
%As can be seen in figure 13, the DouRN model has the fastest growth in win rate at the beginning of the training, and although the rate of increase slows down afterwards, there is no significant trend of convergence. Therefore, A longer training time will result in a higher winning rate model.

\subsection{Evaluation on Bidding System}

For the bidding system, we used a different assessment method. Instead of playing against other models as in Sec. \ref{sec:ern}, we chose 4 human players to play against a model containing the bidding system and calculated the win rate. These 4 players include 3 regular DouDiZhu players in their twenties and 1 top DouDiZhu player in his fifties. They will play against the DouRN network with and without a bidding system, completing a total of 100 games and keeping track of the winning percentage.

\begin{table}
	\centering
	\caption{Winning percentage for human players against DouRN, calculated by the sum of games won by human, divided by the total number of games (50).}
	\begin{tblr}{
			width = \linewidth,
			colspec = {Q[288]Q[298]Q[344]},
			vline{2} = {-}{},
			hline{1,6} = {-}{0.08em},
			hline{2} = {-}{0.05em},
		}
		& With Bidding System & Without Bidding System \\
		Ordinary player1    & 38\%                & 40\%                   \\
		Ordinary player2    & 42\%                & 42\%                   \\
		Ordinary player3    & 32\%                & 42\%                   \\
		Experienced player1 & 46\%                & 52\%                   
	\end{tblr}
\end{table}\label{tab:5}

 When playing against models that lack a bidding system, the distribution of cards and characters to each player is random. Thus, it is possible for a player to receive a disadvantageous hand yet still become the landlord, whatever the player is human or machine. This random allocation ensures fairness in the game, preventing any bias towards the landlord or peasant in terms of winning rates. 

As indicated in Tab. \ref{tab:5}, the introduction of a bidding system has more or less decreased the winning rate of human players. Even experienced players have their winning rates lower than 50\% due to the introduction of the bidding system. And overall, the model can beat all human participants.

%did significantly impact the average player, except for player 3, where there was no notable change in the win rate for the other two players. However, the model incorporating the bidding system demonstrated a substantial performance advantage over the model without it when playing against experienced players, even surpassing the winning rate from being equal to slightly better. 

Based on the aforementioned data, it can be inferred that the inclusion of a bidding system can yield superior results when playing against human opponents. Furthermore, it is plausible to anticipate improved performance against other models as well. 

\section{Conclusion and Future Work}

In this study, we enhanced the DouZero model by introducing the residual neural networks and incorporating a bidding system. Residual neural networks have been shown to be effective on this problem, and the bidding system can be valuable to any other DouDiZhu model.

Despite the advantages of DouRN, there is still room for further optimization. Firstly, additional computational resources are needed to validate and optimize larger neural network. Secondly, the stability of model can possibly be enhanced by incorporating Monte Carlo search techniques. Lastly, we intend to incorporate off-policy learning into the model training process to improve overall efficiency. 

\bibliographystyle{plain} 
\bibliography{ref}

\end{document}